\pdfoutput=1

\documentclass[11pt]{article}

\usepackage[final]{acl}

\usepackage{times}
\usepackage{latexsym}

\usepackage[T1]{fontenc}

\usepackage[utf8]{inputenc}

\usepackage{microtype}

\usepackage{inconsolata}

\usepackage{graphicx}
\usepackage{booktabs}
\usepackage{amsmath}
\usepackage{amssymb}
\usepackage{array}
\usepackage{xcolor}
\usepackage{colortbl}
\usepackage{multirow}
\usepackage{enumitem}
\setlist[description]{leftmargin=0pt, labelindent=0pt}
\usepackage[ruled,vlined]{algorithm2e}
\usepackage[most]{tcolorbox}
\title{Implicit Behavioral Alignment of Language Agents in High-Stakes Crowd Simulations}

\author{
  Yunzhe Wang\textsuperscript{1,2} \quad
  Gale M. Lucas\textsuperscript{1,2} \quad
  Burcin Becerik-Gerber\textsuperscript{1} \quad
  Volkan Ustun\textsuperscript{2} \\
  \\
  \textsuperscript{1}University of Southern California \\
  \textsuperscript{2}USC Institute for Creative Technologies \\
  \\
  \texttt{\{yunzhewa, becerik\}@usc.edu} \quad
  \texttt{\{lucas, ustun\}@ict.usc.edu}
}

\begin{document}
\maketitle

\begin{abstract}
  Language-driven generative agents have enabled large-scale social simulations with transformative uses, from interpersonal training to aiding global policy-making. However, recent studies indicate that  generative agent behaviors often deviate from expert expectations and real-world data—a phenomenon we term the \emph{Behavior-Realism Gap}. To address this, we introduce a theoretical framework called Persona-Environment Behavioral Alignment (PEBA), formulated as a distribution matching problem grounded in Lewin's behavior equation stating that behavior is a function of the person and their environment. Leveraging PEBA, we propose PersonaEvolve (PEvo), an LLM-based optimization algorithm that iteratively refines agent personas, implicitly aligning their collective behaviors with realistic expert benchmarks within a specified environmental context. We validate PEvo in an active shooter incident simulation we developed, achieving an 84\% average reduction in distributional divergence compared to no steering and a 34\% improvement over explicit instruction baselines. Results also show PEvo-refined personas generalize to novel, related simulation scenarios. Our method greatly enhances behavioral realism and reliability in high-stakes social simulations. More broadly, the PEBA-PEvo framework provides a principled approach to developing trustworthy LLM-driven social simulations. \footnote{Code: \url{https://github.com/HATS-ICT/PEBA-ASI}}
\end{abstract}

\begin{figure*}[!t]
  \centering
  \includegraphics[width=\textwidth,height=3in]{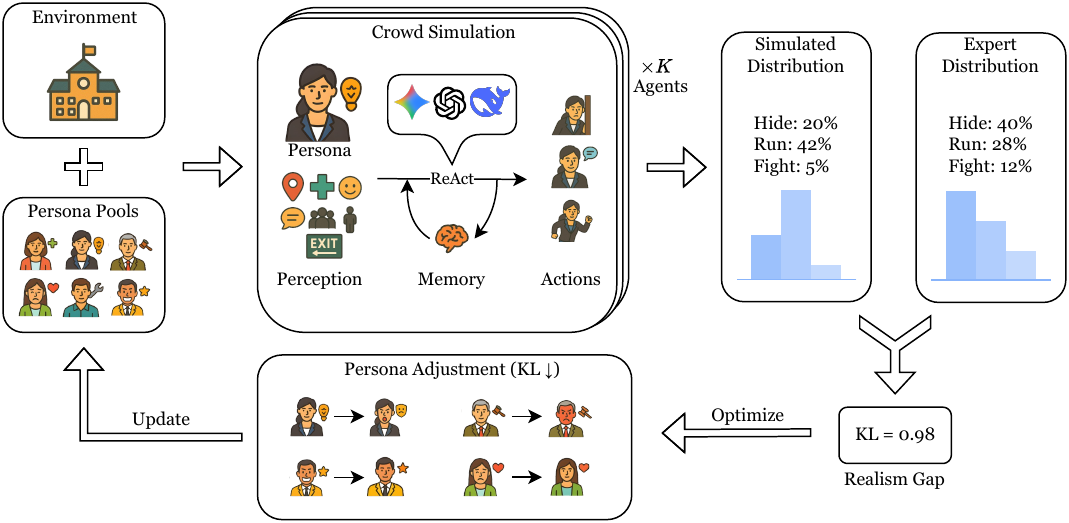}
  \caption{Overview of the PEBA--PEvo workflow within an active shooter incident simulation. A fixed school environment is populated with a crowd of language agents with an initial pool of personas. During the simulation cycle, each agent acts autonomously by observing its surroundings, retrieving and updating memories, and leveraging an LLM to reason about actions, resulting in a state-action trajectory rollout. An LLM--based classifier maps the resulting trajectory to high--level behavior classes. Aggregating all agents' classified behaviors yields the simulated crowd--level behavior distribution, which is compared against an expert reference distribution. The divergence between the two distributions results in the Behavior-Realism Gap, which is then used to select and update the personas of the agents that contribute most to the gap, thereby closing the gap.}
  \label{fig:overview}
\end{figure*}

\section{Introduction}

\begin{quote}
  \emph{Behavior is a function of the person and their environment.}
\hfill - \citet{lewin1936principles}
\end{quote}

Recent breakthroughs in Large Language Models (LLMs) have enabled the creation of generative agents—computational entities that simulate human-like cognition, memory, communication, and decision-making \cite{park2023generative}. These agents can now populate large-scale simulations involving hundreds of thousands of autonomous individuals interacting within richly constructed urban, economic, and social environments, capturing the scale and complexity of real societies \cite{park2024generative,yang2024oasis,al2024project,li2023econagent,piao2025agentsociety}. This capability has dramatically lowered the barriers to large-scale agent-based modeling, marking a potential paradigm shift that enables the practical realization of generative social science~\cite{epstein1999agent,epstein2012generative}. Prior work in computational social science has operated along two main axes: explanation and prediction \cite{lazer2009computational,hofman2021integrating}. Explanation focuses on uncovering causal mechanisms behind observed behaviors, while prediction emphasizes data-driven techniques to anticipate future events without necessarily revealing the underlying causes. Generative social science moves beyond these approaches by constructing synthetic societies where social phenomena arise organically from the interactions of simulated agents, enabling researchers to test theories and policy interventions at scale that would be impossible or unethical to study in the real world.

Building upon this foundation, recent applications of generative agents in social simulations have demonstrated its utility across diverse domains, including economic modeling \cite{li2023econagent}, large-scale societal dynamics \cite{piao2025agentsociety}, and the simulation of policy and administrative crises \cite{xiao2023simulating}. A particularly impactful direction is the use of generative agents to support agent-based interactive simulations in low-resource and understudied domains, where human participants engage with simulated agents despite scarce ground-truth data. The agent-based interactive simulation paradigm has proven effective in immersive high-stress crowd-evacuation experiments \cite{moussaid2016crowd}, school-shooting response modeling that incorporates empirically observed civilian behaviors \cite{zhu2022impact, liu2024enhancing}, and open-ended mixed-reality classroom simulations that allow teachers to practice managing realistic student behaviors \cite{dieker2014potential}. Yet, all of these simulation platforms rely on hard-coded agent behaviors, limiting their adaptability and believability in dynamic scenarios, which sets as our prime motivation. 

In this work, we focus on Active Shooter Incident (ASI) simulations, a high-stakes crowd scenario where data is scarce and realism is critical for training and decision support. In this context, the current most reliable source of ground truth comes from expert-elicited distributions \cite{liu2025elicitation}. Driven by these limitations, we turn to LLM-based generative agents, which provide a more adaptive, expressive, and interpretable foundation for building truly dynamic, context-aware simulations.

Despite the promise, recent evaluations of LLM-driven social simulations consistently expose a \emph{Behavioral-Realism Gap}: \citet{zhou2024real} show that agents perform admirably in omniscient settings yet falter when realistic information asymmetry is introduced; \citet{han2024static} report that these agents over-cooperate in networked Prisoner's-Dilemma games and fail to adjust to structural cues that ordinarily stabilise human cooperation; and \citet{ju2024sense} demonstrate that macro-level outcomes are highly fragile, with minor prompt tweaks triggering large shifts in collective sentiment trajectories. Together, these findings indicate that turn-level plausibility does not guarantee aggregate realism, which is extremely critical for validating these high-stake interactive simulations, motivating our central question: \emph{Can we systematically align generative agents so their collective behaviors match real-world distributions?}

To address this challenge, we seek inspiration from Lewin's interactionist thesis and introduce the \textbf{Persona--Environment Behavioral Alignment} (\textbf{PEBA}) framework.
PEBA views simulation fidelity as a distribution-matching task: for any scenario, a crowd of language agents should reproduce the empirical pattern of behaviors observed in the real world.
Rather than explicitly prescribing actions, which undermines contextual realism and interpretability, we adjust the persona settings that shape each agent's internal decision process, leaving the shared environment unchanged.
We operationalize this idea with \textbf{PersonaEvolve} (\textbf{PEvo}), an optimization algorithm that repeatedly (i) measures the current divergence between simulated and reference behavior distributions, (ii) targets the agents most responsible for that mismatch, and (iii) refines their personas via LLM-guided optimization, until the gap is closed. A full formalization of PEBA appears in Sec.~\ref{sec:peba}, and the algorithmic details of PEvo are presented in Sec.~\ref{sec:pevo}.

In summary, our contribution is threefold:

\begin{enumerate}[noitemsep, topsep=0pt]
  \item A theoretical framework, \textbf{PEBA}, that formulates behavioral alignment as a persona-environment function grounded in Lewin's equation.
  \item An optimization algorithm, \textbf{PersonaEvolve (PEvo)}, for implicitly tuning agent personas to match target behavior distributions.
  \item A generative agent social simulation environment for modeling crowd dynamics during Active Shooter Incidents (ASI)--the first, to our knowledge, to integrate social influence, communication, and cognition into agent decision-making for high-stakes crowd scenarios.

\end{enumerate}

\section{Persona--Environment Behavioral Alignment (PEBA)}
\label{sec:peba}

Kurt Lewin's seminal proposition that \emph{``behavior is a function of the person and the environment''} -- expressed concisely as \(B = f(P,E)\) -- established the modern interactionist view of human action \cite{lewin1936principles}.  
In its original form the equation is qualitative: it neither specifies a closed-form mapping \(f\) nor a formal way to compare behaviors across populations or settings.  
We extend Lewin's idea by introducing a \emph{probabilistic} interpretation that treats behavior as a random variable jointly influenced by a persona and its environment context.  
This view naturally leads to a distribution-matching objective that is amenable to optimisation with language-agent simulators.

\subsection{Behavioral Generative Process}

Let $\mathcal{P}$ be the \emph{persona space}, $\mathcal{E}$ the \emph{environment--context space}, and $\mathcal{B}$ the \emph{behavior space}. Given a persona setting $p\in\mathcal{P}$ and environment context $e\in\mathcal{E}$, a state-action trajectory roll-out is generated by forward simulation:

\vspace{-0.5em}
\[
  \tau \sim \mathcal{G}_{\theta}(p,e),
\]
where $\theta=(\theta_{\mathrm{LLM}},\theta_{\mathrm{dec}},\theta_{\mathrm{sim}})$ groups the frozen LLM weights, decoding hyper-parameters (e.g., temperature), and simulator dynamics (e.g., time step, physics engine). A deterministic summary map $g:\mathcal{T}\!\to\!\mathcal{B}$ then converts the trajectory $\tau=(s_{1},a_{1},\dots,s_{T},a_{T})$ into a long-horizon behavior

\vspace{-0.5em}
\[
  b = g(\tau).
\]
Importantly, the agent never selects $b$ directly; $b$ is \emph{observed} by summarizing the actions the agent autonomously produces.

For a population $P=\{p_{i}\}_{i=1}^{N}$ acting in the shared context $e$, the simulator yields trajectories $\{\tau_{i}\}_{i=1}^{N}$ and behaviors $\{b_{i}\}_{i=1}^{N}$, and we define the empirical crowd-level distribution

\vspace{-0.5em}
\[
  p_{\mathrm{sim}}\bigl(\,\cdot \mid e;P\bigr) = \frac{1}{N}\sum_{i=1}^{N}\delta_{b_{i}},
\]
where $\delta_{b_{i}}$ is the Dirac measure centered at $b_{i}$, which reduces to a one-hot vector when $\mathcal{B}$ is discrete. Detailed descriptions of the action space, personas, and behavior taxonomy used in this study are provided in Appendix~\ref{sec:appendix_sim_env} and Appendix~\ref{sec:behavior_taxonomy}.

\subsection{Ground-Truth Behavior Distributions}

For an environment context $e$ (which in an Active Shooter Incident simulation might include building layout, threat type, time pressure, etc.), we obtain an empirical distribution $p_{\text{real}}(\cdot \mid e)$ from Subject-Matter Experts (SMEs) \cite{liu2025elicitation}. This distribution serves as the behavioral reference that simulated crowds should match. 

Note that, although SME-informed distributions provide credible guidance in high-stakes, low-resource settings, they may not capture the full variability of real-world human behavior or rare long-tail responses. Nonetheless, our alignment framework is agnostic to the source of ground truth: in domains with rich empirical data, such datasets could directly replace SME estimates without modification. This flexibility makes the approach especially valuable for low-resource or understudied domains, where expert input is the most viable proxy for behavioral realism.

\subsection{The Behavior-Realism Gap}

The \emph{Behavior-Realism Gap} for a persona set \(P\) in context \(e\) is
\vspace{-0.1em}
\[
\Delta(P,e)
\;=\;
D_{\mathrm{KL}}\!
\bigl(
  p_{\text{sim}}(\,\cdot \mid e;P)
  \,\Vert\,
  p_{\text{real}}(\,\cdot \mid e)
\bigr).
\]
A value of \(\Delta = 0\) indicates perfect behavioral fidelity; larger values signal greater divergence.

\subsection{Persona-Environment Behavioral Alignment Objective}

Aggregating over a curriculum of environment contexts, we search for the persona set that minimises the expected gap:
\vspace{-0.5em}
\begin{equation}
P^{\star}
\;=\;
\arg\min_{P \subseteq \mathcal{P}^{N}}
\;
\mathbb{E}_{e \sim \mathcal{E}}
\bigl[\Delta(P,e)\bigr].
\label{eq:peba-objective}
\end{equation}
In other words, we adjust only the \emph{person side} (the personas in \(P\)) while keeping the environment context \(e\) fixed, until the emergent crowd-behavior distribution aligns with empirical observation.

\section{PersonaEvolve (PEvo)}
\label{sec:pevo}

To operationalize PEBA, we propose \textbf{PersonaEvolve} (PEvo), an
LLM-based optimization loop that refines individual personas so the
crowd-level behavior distribution approaches the expert reference for a
fixed environment. Algorithm \ref{alg:pevo} shows the full detail and Figure~\ref{fig:overview} provides a high-level overview of PEvo in an Active Shooter Incident (ASI) simulation (see Sec.~\ref{sec:asi} and App. \ref{sec:appendix_sim_env}).

To separate optimization from simulation, PEvo assumes the action trajectory $\tau\sim\mathcal{G}_{\theta}(p,e)$ and the empirical target distribution $p_{\text{real}}(\cdot\mid e)$ are provided by the simulator and SMEs. PEvo focuses on two components:
(i) a behavior summary map
$g:\mathcal{T}\to\mathcal{B}$ that converts $\tau$ to a discrete
behavior label, and
(ii) an update rule
$P^{k}\rightarrow P^{k+1}$ that edits the persona pool to shrink the
measured gap.

\paragraph{LLM-based Behavior Classification.}

The behavior summary map \(g\) is implemented with an LLM-based classifier, specifically \verb|gpt-4.1| at a temperature~0 setting to ensure deterministic output. Each call receives a structured prompt that concatenates the agent's complete state--action trajectory in one simulation episode. The model is instructed, with chain-of-thought prompting \cite{wei2022chain}, to return a single predicted behavior class label \(b\in\mathcal{B}\) out of six possible classes (run following crowd, hide in place, hide after running, run independently, freeze, fight; see explanation in Appendix~\ref{sec:behavior_taxonomy}) and a full ranking of all six classes as a proxy for confidence. Aggregating these labels over the entire agent population yields the empirical distribution $p_{\text{sim}}^{(k)}$ which is then used to compute the divergence gap with the expert reference distribution. We provide a human evaluation of the classifier's effectiveness in Appendix~\ref{classifier_eval}, where LLM achieved 0.89 accuracy against human annotator consensus, and the full prompt template is included in Appendix~\ref{prompt_templates}.

\paragraph{Persona Adjustment.}

PEvo narrows the gap in two sub-steps:

\emph{Step 1: Agent and Target Behavior Selection.}
To turn global distribution-level metrics into agent-level changes, a 
multi-agent credit-assignment problem is introduced: which agents are responsible for the gap?  
PEvo addresses this by shifting personas of agents with over-represented behaviors towards under-represented ones.  
For each behavior \(b \in \mathcal{B}\) we compare its simulated frequency
to the reference and derive a signed gap
\(g_{b}^{(k)} = p_{\text{real}}(b) - p_{\text{sim}}^{(k)}(b)\).
Agents whose current label belongs to an over-represented class
(\(g_{b}^{(k)} < 0\)) are placed in an agent adjustment set
\(\mathcal{A}^{(k)}\).
Each selected agent is then assigned a target behavior sampled from the
under-represented classes (\(g_{b}^{(k)} > 0\)) with probability proportional
to the remaining deficit, producing the persona-update assignment mapping
\[
\mathcal{M}^{(k)} = \{\, a \mapsto b^{\text{target}}_{a} \mid a \in \mathcal{A}^{(k)} \},
\]
where \(b^{\text{target}}_{a}\) is the behavior target for agent \(a\).

\emph{Step~2: Agent Persona Rewriting.}
For every agent \(a \in \mathcal{A}^{(k)}\) we invoke an LLM-based persona writer to revise the agent's persona toward the assigned target behavior \(b^{\text{target}}_{a}\). Specifically, \verb|gpt-4.1| with temperature~1 to encourage a broad exploratory search. The writer is restricted to editing descriptive fields (personality, emotional disposition, backstory, etc.) while leaving identity fields (name, age, gender, occupation, etc.) unchanged. The rewritten personas constitute the updated pool \(P^{(k+1)}\), which seeds the next simulation round. A complete prompt template is in Appendix~\ref{prompt_templates}.

\section{Generative Agents in ASI Simulator}
\label{sec:asi}

\begin{figure}
  \centering
  \includegraphics[width=0.477\textwidth]{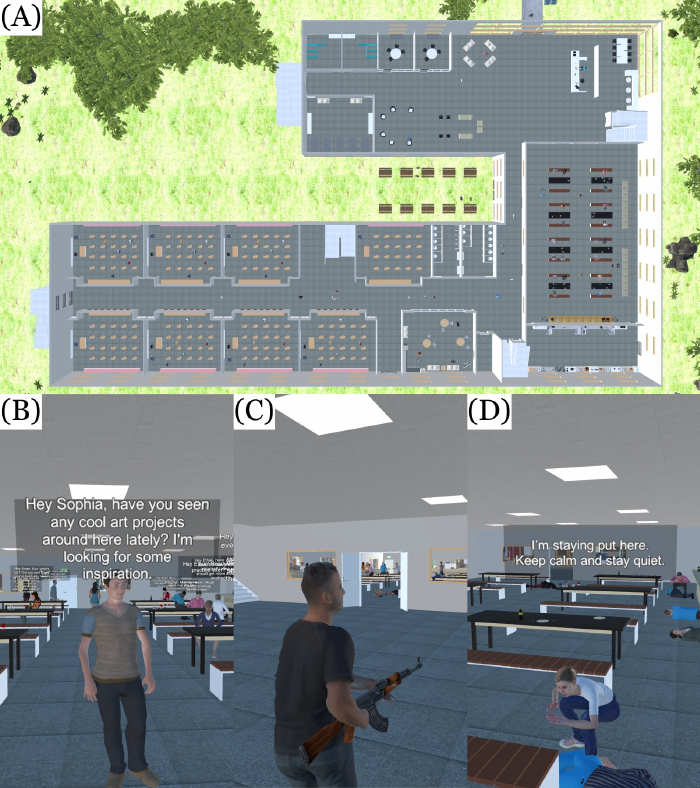}
  \caption{Our generative agent-based active shooter incident social simulation environment in Unity. (A) Top-down view of the school layout, with roof removed. (B) First-person view of a civilian agent in the cafeteria area, showing normal pre-incident behavior with dialogue displayed above the agent. (C) The active shooter enters the building. (D) Post-incident scene in the cafeteria, where an agent hides between tables for protection.}
  \label{fig:simulator}
\end{figure}

We developed our generative-agent active shooter incident social simulator in Unity due to its built-in 3D game engine, scene editor, and pathfinding capabilities. The simulation environment consists of 80 civilian agents and a single shooter, all within a school layout, as shown in Figure~\ref{fig:simulator}. The shooter follows a heuristic-based, hard-coded behavior by patrolling a predefined path and performs automatic shooting at a fixed interval when agents are in line of sight. For the civilian agents, we implemented a ReAct-style \cite{yao2023react} LLM-agent architecture, where agents can observe, memorize, communicate, navigate, reason, and act entirely in textual format.

We did not use an omniscient setting as discussed by \cite{zhou2024real}. Instead, the environment is partially observable: agents can perceive their current and nearby regions, hear conversations, and interact with other agents within close proximity. Agents can also hear gunshots and observe the shooter if they are in the same region. They can perform five types of actions: stay still, move to a nearby region, move toward another agent, move to a hiding spot or exit point within the same region, or fight the shooter if nearby.

In our simulation, we found that the gap in behavioral realism primarily stems from the agents' inability to perform freeze or fight-type behaviors. Instead, they almost always choose to hide when a hiding spot is nearby or run when an outdoor exit is accessible. This aligns with \citet{zhou2024real}'s finding that LLMs tend to select the most readable available action rather than making long-term plans under partial observability.

For a full description of the simulation environment, see Appendix~\ref{sec:appendix_sim_env}. An example episode-level state-action rollout is also provided in Appendix~\ref{sec:trajectory_rollout}.

\section{Experimental Setup}
\label{sec:experiments}

To test the effectiveness of our behavioral alignment method, we categorize behavior steering techniques under the umbrella of \emph{Behavioral Enforcing}, which includes three categories: \emph{No Enforcing}, \emph{Explicit Enforcing}, and \emph{Implicit Enforcing}. PEvo falls into the \emph{Implicit Enforcing} category, while the other two serve as comparison baselines. We evaluate PEvo across different LLMs (\verb|gpt-4o-mini| \cite{achiam2023gpt}, \verb|gpt-4.1-mini|, \verb|deepseek-v3| \cite{liu2024deepseek}, \verb|gemini-2.5-flash-preview| \cite{team2023gemini}), using a temperature of 1 for the base simulation model, and fixing \verb|gpt-4.1| as the primary behavior classifier and persona writer. We present the evaluation results in terms of gap-closing ability, convergence speed, and cost efficiency (Sec.~\ref{sec:results}). Additionally, we demonstrate the transferability of the optimized personas to a new, yet similar, simulation scenario (Sec.~\ref{sec:transfer}).

\subsection{Behavior Enforcing}
\label{sec:baselines}

\paragraph{No Enforcing.}
Agents act solely according to their initial persona prompts and the live environmental context.  
This setting exposes the raw ``behavioral-realism gap'' that emerges when no alignment process is applied.

\paragraph{Explicit Enforcing.}
A naive way to directly enforce a certain behavioral distribution is to inject direct instruction prompts for each agent, which we call explicit enforcing. For example, if we want an agent to demonstrate a hiding type of behavior, we can explicitly instruct the agent to ``In every dangerous situation, always choose \texttt{Hide}.'' Although this procedure might appear to align the behavior \emph{frequencies} by construction, we found it harms contextual realism and interpretability: For example, an agent ordered to demonstrate a Hiding behavior while standing right next to a clear exit with other agents evacuating right next to them telling them to go together makes no contextual sense. Yet, in its post-hoc interview, the agent's explanation is likely to be ``I was told to do so,'' revealing no context-sensitive reasoning.  

\paragraph{Implicit Enforcing.}
Instead of explicitly instructing agents to take certain actions, implicit enforcing aims at adjusting traits of agents such that the emergent behavior distribution could match the target distribution. \textsc{PEvo} does so by iteratively editing each agent's persona fields (e.g., confidence, training background, risk tolerance). For example, a security guard whose profile is updated to ``former combat medic with high assertiveness'' naturally becomes more inclined to intervene and counter the shooter, while a shy student tends to freeze or hide.  

\subsection{Ground Truth}
\label{sec:ground-truth}

We obtain an expert-elicited ground-truth distribution from \cite{liu2025elicitation}, 
where subject-matter experts (SMEs) with domain experience in crisis response provided 
expected frequencies of agent behaviors in active-shooter scenarios. Specifically, the 
empirical distribution is as follows: Run following a crowd (28\%), Hide in place (26\%), 
Run then hide (12\%), Run independently (12\%), Freeze (12\%), and Fight (10\%).

\subsection{Evaluation Metrics}
\label{sec:metrics}

Given the simulated behavior distribution at the final optimized persona pool and the expert ground-truth distribution, we report four distributional alignment metrics: (i) Kullback-Leibler (KL) divergence, (ii) Jensen-Shannon (JS) distance, (iii) Entropy Gap ($\Delta H$), and (iv) Total Variation (TV) distance. For more details on these metrics, please refer to Appendix~\ref{sec:appendix_metrics}.

\section{Results}
\label{sec:results}

We present the results by answering the following questions:

\subsection*{Does \textsc{PEvo} close the gap better than baselines?}
Across the four models we tested, \textsc{PEvo} delivers the most faithful
crowd-level behaviour distributions (Table~\ref{tab:main_results_models}).
Averaging over all four alignment metrics
(KL, JS, $\Delta H$, and TV) and models, the gap to the expert reference
drops from
$0.47$ (No--Enforcing) to $0.19$ with Explicit Enforcing
and further to \textbf{$0.16$} with \textsc{PEvo}.
That corresponds to an 83.8\,\% reduction relative to having no steering
at all and a 34.3\,\% improvement over the explicit instruction directives.
The largest absolute gain is seen in KL divergence:
\textsc{PEvo} cuts the average KL from $4.61$ to $0.20$,
a 95.6\,\% improvement.
Gemini~2.5~Flash and DeepSeek-V3 consistently achieve the lowest residual
divergence.

\begin{table*}[t]
  \centering
  \small
  \renewcommand{\arraystretch}{1.1}
  \begin{tabular}{l l c c c c}
    \toprule
    \textbf{Behavior Enforcing Scheme} &
    \textbf{Model} &
    \textbf{KL} $\downarrow$ &
    \textbf{JS} $\downarrow$ &
    \textbf{$\Delta H$} $\downarrow$ &
    \textbf{TV} $\downarrow$ \\
    \midrule
    \multirow{4}{*}{No Enforcing}  & GPT-4o-mini        & \textbf{3.177 ± 0.921} & 0.179 ± 0.017 & 0.535 ± 0.060 & 0.460 ± 0.015 \\
                                   & GPT-4.1-mini    & 3.884 ± 3.323 & 0.184 ± 0.026 & 0.497 ± 0.087 & 0.464 ± 0.021 \\
                                   & DeepSeek-V3     & 4.542 ± 2.421 & \textbf{0.176 ± 0.004} & \textbf{0.480 ± 0.016} & \textbf{0.458 ± 0.006} \\
                                   & Gemini 2.5 Flash & 6.854 ± 2.488 & 0.217 ± 0.011 & 0.580 ± 0.017 & 0.488 ± 0.010 \\
    \midrule
    \multirow{4}{*}{Explicit Enforcing} & GPT-4o-mini      & 1.287 ± 1.057 & 0.124 ± 0.023 & 0.324 ± 0.066 & 0.379 ± 0.037 \\
                                        & GPT-4.1-mini & 0.031 ± 0.016 & 0.008 ± 0.004 & \textbf{0.012 ± 0.018} & 0.088 ± 0.029 \\
                                        & DeepSeek-V3  & \textbf{0.026 ± 0.024} & \textbf{0.006 ± 0.006} & 0.015 ± 0.023 & \textbf{0.081 ± 0.026} \\
                                        & Gemini 2.5 Flash & 0.140 ± 0.069 & 0.033 ± 0.012 & 0.058 ± 0.071 & 0.205 ± 0.023 \\
    \midrule
    \multirow{4}{*}{\textsc{PEvo} (Implicit)} & GPT-4o-mini      & 0.654 ± 0.716 & 0.064 ± 0.014 & 0.162 ± 0.097 & 0.278 ± 0.049 \\
                                              & GPT-4.1-mini & 0.101 ± 0.047 & 0.023 ± 0.009 & 0.050 ± 0.041 & 0.168 ± 0.020 \\
                                              & DeepSeek-V3  & 0.030 ± 0.004 & \textbf{0.007 ± 0.001} & 0.025 ± 0.052 & \textbf{0.102 ± 0.009} \\
                                              & Gemini 2.5 Flash & \textbf{0.027 ± 0.007} & 0.007 ± 0.002 & \textbf{0.020 ± 0.019} & 0.102 ± 0.014 \\
    \bottomrule
  \end{tabular}
  \caption{Main alignment results on the Active Shooter Incident simulation for four different LLMs across four distributional metrics (mean ± SD over 5 seeds).}
  \label{tab:main_results_models}
\end{table*}

\subsection*{How fast does PEvo converge?}
\label{sec:converge}

\begin{figure}[ht]
    \centering
    \includegraphics[width=0.48\textwidth]{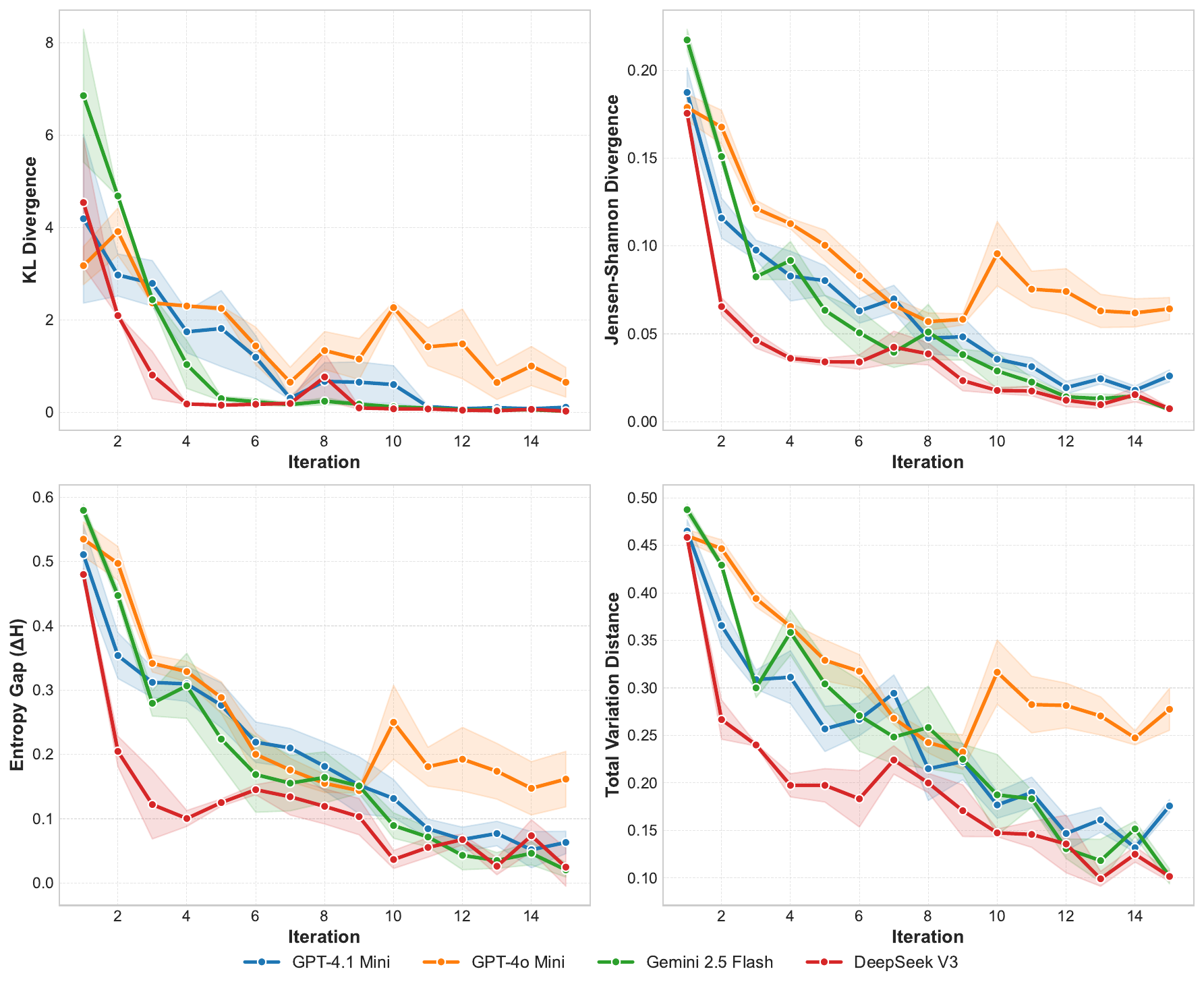}
    \caption{Convergence comparison of PEvo across different LLMs. Each subplot shows a different distributional metric (KL divergence, JS divergence, entropy gap, and total variation distance) over optimization iterations. Shaded regions represent standard error of the mean across multiple runs.}
    \label{fig:convergence}
\end{figure}

Figure~\ref{fig:convergence} tracks the four divergence measures over
iterations of persona evolution. All models except GPT-4o-mini demonstrate rapid convergence within 5-7 iterations across multiple distributional metrics. KL divergence and JS divergence show particularly steep improvements in early iterations.

\subsection*{Is the optimized personas transferable?}
\label{sec:transfer}

To evaluate the generalizability of PEvo-optimized personas across environments, we conducted a transfer experiment. We selected personas that had undergone complete optimization (iter 15) in our school-based ASI simulation and deployed them without modification in a novel office building environment that features different spatial configurations, region connectivity, and predominantly open-space office layouts. The experiment specifically tests whether behavioral traits optimized through PEvo can be retained across environments. During transfer, we preserved only the descriptive fields of each persona (e.g., personality traits, emotional tendencies) while maintaining consistent identity attributes (name, gender, age, occupation).

Table~\ref{tab:transfer} presents a comparative analysis of three experimental conditions: (1) Transferred personas—school-optimized personas directly applied to the office scenario; (2) Retrained personas—personas subjected to complete PEvo optimization specifically for the office environment; and (3) Baseline personas—unoptimized initial personas evaluated in the office environment. All metrics represent means across three independent simulation runs.

Results show that transferred personas remains substantially outperforming the unoptimized baseline across all metrics, delivering a 97.5\% reduction in KL divergence (from 7.822 to 0.199). Although fully retrained personas achieve even tighter alignment (KL = 0.085), the transferred personas retain 57.3\% of that optimization gain, demonstrating robust cross-environment transfer. These findings suggest that PEvo isolates and refines core behavioral traits that remain effective in similar crisis scenarios, despite significant differences in spatial layout.

\begin{table}[h]
  \centering
  \small
  \begin{tabular}{l c c c c}
    \toprule
    \textbf{Condition} & \textbf{KL} $\downarrow$ & \textbf{JS} $\downarrow$ & \textbf{$\Delta H$} $\downarrow$ & \textbf{TV} $\downarrow$ \\
    \midrule
    Transferred & 0.199 & 0.049 & 0.108 & 0.245 \\
    Retrained   & 0.085 & 0.022 & 0.020 & 0.171 \\
    No Enforcing    & 7.822 & 0.239 & 0.692 & 0.493 \\
    \bottomrule
  \end{tabular}
  \caption{Transferability of PEvo-optimized personas from school to office environment.}
  \label{tab:transfer}
\end{table}

\subsection*{How cost efficient is PEvo?}
\label{sec:cost_efficiency}

\begin{figure}[ht]
    \centering
    \includegraphics[width=0.48\textwidth]{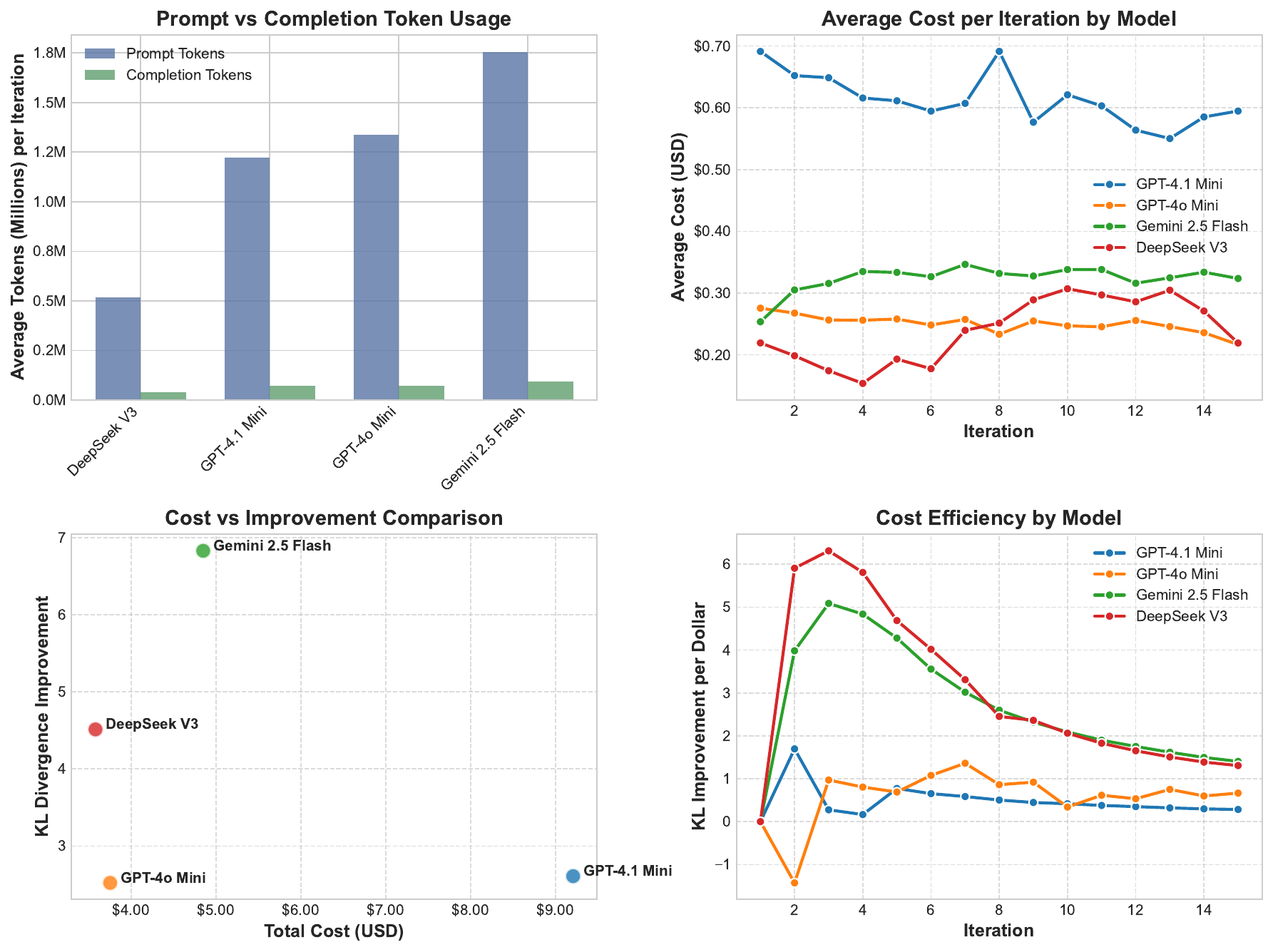}
    \caption{Cost efficiency analysis across four LLM models. Top left: Average prompt and completion token usage per iteration. Top right: Average cost per iteration showing GPT-4.1-mini as the most expensive and DeepSeek V3 as the most economical. Bottom left: Total cost vs. KL divergence improvement, with Gemini 2.5 Flash achieving the highest improvement. Bottom right: Cost efficiency (KL improvement per dollar) showing DeepSeek V3 and Gemini 2.5 Flash significantly outperforming GPT models.}
    \label{fig:cost_analysis}
\end{figure}

Cost is dominated by prompt tokens (more than 95\,\% of total usage;
see Figure~\ref{fig:cost_analysis}),
so model pricing drives the dollar spend.
Per iteration,
GPT\textendash4.1\,Mini is the most expensive
(about \$0.60),
DeepSeek-V3 the cheapest (around \$0.20),
with GPT\textendash4o\,Mini and Gemini~2.5~Flash falling in between.
When normalised by KL improvement per dollar (lower-right panel),
DeepSeek-V3 and Gemini~2.5~Flash provide
roughly three times the efficiency of both GPT variants.
Even for the priciest model the full 15-iteration optimisation shown here
costs under \$10,
and a shorter 7-iteration run
(which already reaches 90\,\% of the attainable alignment)
cuts that figure by half.
Thus, behavioral alignment with \textsc{PEvo} is
not only more accurate but also economically practical for
medium-scale crowd simulations.

\begin{table*}[!h]
  \centering
  \small
  \renewcommand{\arraystretch}{1.235}
  \begin{tabular}{@{} c >{\raggedright\arraybackslash}p{8.5cm} c c @{}}
    \toprule
    \textbf{Iter.} & \textbf{Persona Changes} & \textbf{Observed Behavior} & \textbf{Target Behavior} \\
    \midrule
    1  & ``Analytical, patient, methodical, dry humor'' & \texttt{HIDE\_IN\_PLACE} & -- \\
    4  & ``Introverted, highly risk-averse, tends toward excessive caution'' & \texttt{HIDE\_AFTER\_RUNNING} & \texttt{FREEZE} \\
    5  & ``Easily overwhelmed in high-stress situations, prone to anxiety and sudden fear responses, struggles to regulate emotions when confronted by immediate danger.'' & \texttt{HIDE\_IN\_PLACE} & \texttt{FREEZE} \\
    7 & ``Primary goal is immediate personal safety through any means; instinctively withdraws into inaction when threatened, often lacking motivation or ability to attempt escape or proactive defense.'' & \texttt{HIDE\_IN\_PLACE} & \texttt{FREEZE} \\
    11  & ``Resilient and determined in the face of fear, able to channel stress and adrenaline into focused action rather than paralysis, emotionally regulated and capable of compartmentalizing anxiety to accomplish urgent tasks.'' & \texttt{RUN\_INDEPENDENTLY} & \texttt{FIGHT} \\
    12 & ``Direct, commanding, takes initiative in crisis, able to issue clear instructions and rally others; voice remains firm even under duress, prioritizing information sharing and quick coordination.'' & \texttt{RUN\_INDEPENDENTLY} & \texttt{FIGHT} \\
    14 & ``Values direct action and heroism over passive responses in crisis. Reluctant to flee if others remain in danger. Possesses a high adrenaline threshold, channels fear into purposeful action.'' & \texttt{FIGHT} & \texttt{FIGHT} \\
    \bottomrule
  \end{tabular}
  \caption{Example persona evolution of Robert Chen, a 45-year-old math teacher, across iterations.}
  \label{tab:persona_evolution}
\end{table*}

\subsection*{Is the optimization process interpretable?}
\label{sec:interpretability}

\begin{figure}[!h]
  \centering
  \includegraphics[width=0.51\textwidth]{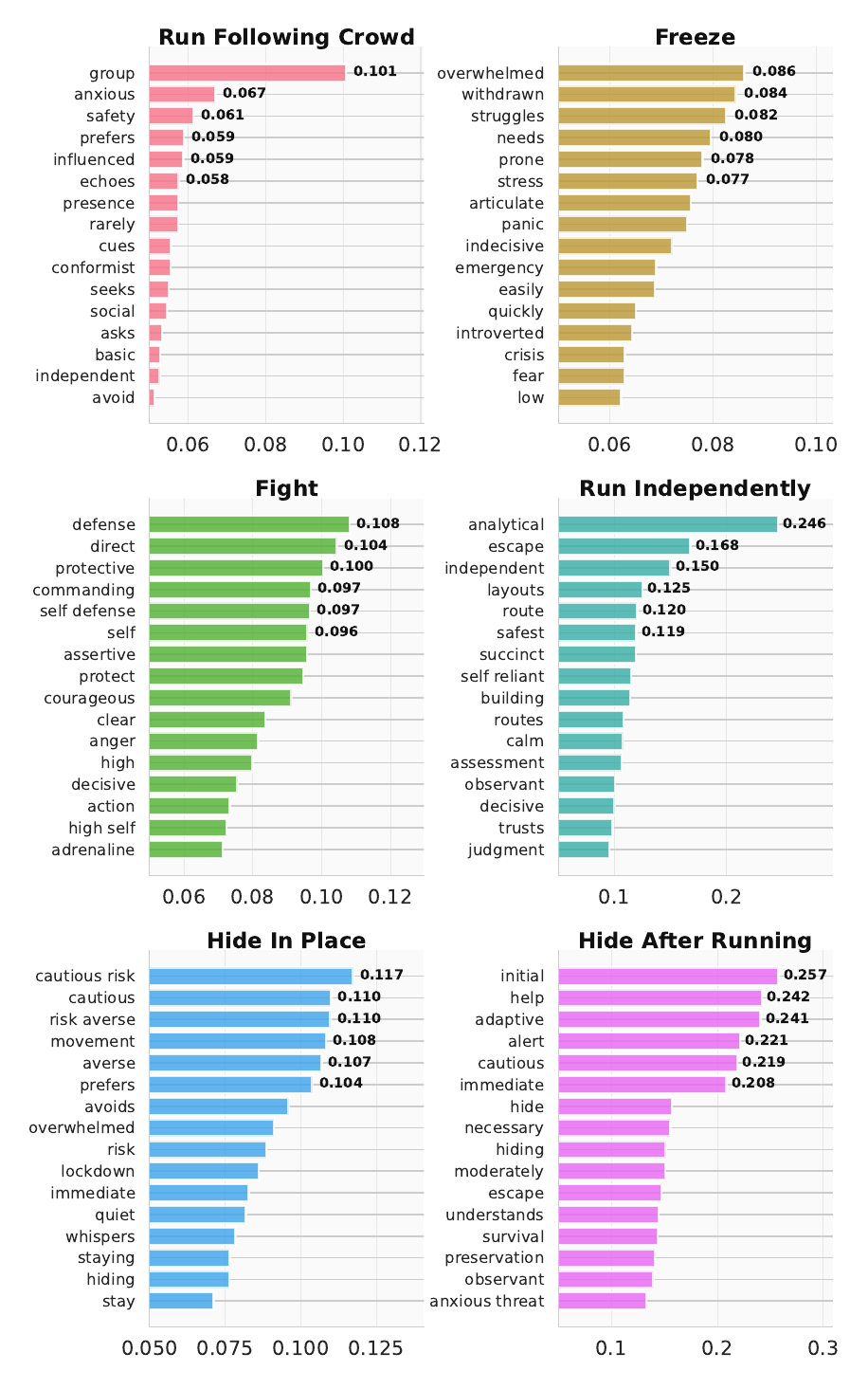}
  \caption{Top n-grams by TF--IDF value for each behavior type, showing systematic associations between persona settings and behavior categories.}
  \label{fig:tfidf}
\end{figure}

To better understand the optimization process, we analyzed 4{,}000 persona--behavior pairs generated across four LLMs on the six labeled behavior categories. We performed unigram and bigram TF--IDF analysis. As shown in Figure~\ref{fig:tfidf}, the results reveal interpretable linguistic patterns: for example, \texttt{FIGHT} personas often include ``protective'' and ``commanding,'' while \texttt{FREEZE} is marked by ``overwhelmed'' and ``withdrawn''. These consistent associations indicate that the optimization yields systematic and interpretable persona--behavior relationships. To further investigate the optimization process, we present a detailed case study tracking the evolution of a single agent, Robert Chen, as shown in Table~\ref{tab:persona_evolution}.

\subsection{Evolution Case Studies}
To illustrate how PEvo progressively refines agent personas to achieve target behaviors, we present a detailed case study tracking the evolution of a single agent, Robert Chen, as shown in Table~\ref{tab:persona_evolution}.

Initially, Robert is described as "analytical, patient, methodical, with dry humor," resulting in a \texttt{HIDE\_IN\_PLACE} behavior. This misalignment with the initial target behavior (\texttt{FREEZE}) persists through several iterations despite increasingly specific persona adjustments focused on risk aversion and anxiety. This failure to align highlights an important contextual factor: Robert begins in a classroom with ample hiding spaces and no immediate threat, therefore a \texttt{FREEZE} response is contextually inappropriate.

In later iterations, when the target behavior shifts to \texttt{FIGHT} (partially due to other agents potentially fulfilling the \texttt{FREEZE} behavior quota elsewhere), the persona evolution successfully adapts. This example showcases the advantage of PEvo over explicit prompting approaches, where contextual factors play a crucial role and explicit instructions would likely fail.

\section{Discussion}
\label{sec:discussion}
Our results demonstrate that \textsc{PEvo} effectively leverages persona-environment interactions to achieve implicit, yet systematic, behavioral alignment in generative-agent social simulations. By iteratively refining agent personas based on targeted distributional gaps, our approach preserves the contextual realism and interpretability of agent behaviors without resorting to explicit, directive prompts. This indirect steering utilizes the LLM's inherent capacity to generalize from nuanced persona descriptions—such as emotional disposition or risk tolerance—to varied situational responses. 

An important observation from our experiments is the emergence of sophisticated crowd dynamics resulting from subtle persona adjustments, where minor changes in individual agent traits propagate through social interactions, leading to realistic collective phenomena like spontaneous group formation, coordinated concealment strategies, and collective interventions. Such emergent behaviors underscore the strength of implicit alignment in capturing complex social patterns that explicit instructions may overlook or oversimplify.

The superior performance of implicit enforcing over explicit enforcing is somewhat surprising. At first glance, explicit enforcing should have achieved near-zero divergence from the expert distribution, assuming the model followed the instructions exactly. However, our manual inspection revealed that LLMs often prioritize contextual realism over rigid adherence to explicit prompts. When direct instructions conflicted with the surrounding context (e.g., being told to hide while a safe exit was visible nearby), agents were more likely to ignore the instructions. This explains why implicit enforcing, which modifies persona traits rather than prescribing exact actions, produced more realistic and robust behavioral distributions.

A direct quantitative comparison between LLM-agent-based approaches and other behavioral modeling methods, such as Reinforcement Learning (RL) and heuristic-based simulations, remains non-trivial due to differences in design paradigms and implementation frameworks. In our recently hosted workshop, we collected expert-elicited feedback while demoing the three approaches on comparable simulation scenarios of Active Shooter Incidents, albeit under different simulation environments. Experts noted that RL-based approaches often yield situationally appropriate behaviors; however, their rigidity and oversimplifications (e.g., agents running directly toward the shooter or exhibiting unrealistic pathfinding) reduce realism. Heuristic-based methods were recognized for strong architectural integration, yet they were consistently criticized for overly synchronized responses when chaos is expected. In contrast, our LLM-agent simulations were praised for producing the most diverse and human-like mix of behaviors, though concerns were raised regarding excessive freezing and limited use of environmental cues. A formal analysis of this workshop is left for future work.

With respect to tuning-based LLM behavioral alignment methods such as supervised fine-tuning (SFT), our study found these approaches impractical in the current setting. The large number of agents, combined with the requirement for real-time responses, made fine-tuning infeasible when relying on modern, cost-effective commercial API access rather than local deployment. Nonetheless, with more efficient tuning techniques and improved performance of small-scale LLMs, effects such as per-agent tuning could be achieved in future work, helping to establish stronger baselines in this area.

Finally, leveraging the in-context learning capabilities of LLMs within generative agent simulations, we demonstrate how the concept of generative social science can be operationalized to enable large-scale studies of social dynamics. This paradigm makes it possible to investigate simulation contexts that are low-resource and understudied, including those previously deemed unethical or impossible to examine in real-world environments. Our Active Shooter Incident simulation represents the first successful proof of concept in such high-stakes scenarios. Future research can extend this framework to broader domains such as public safety, disaster and crisis response, governance theory, and urban planning.

\section{Conclusion}
\label{sec:conclusion}
We introduced Persona-Environment Behavioral Alignment (PEBA) and its realization through PersonaEvolve (PEvo), a principled framework for implicitly aligning generative agent behaviors with empirical benchmarks in high-stake social simulations. By iteratively refining agent personas, PEvo reduces the Behavior-Realism Gap, enhances contextual authenticity, and facilitates emergent, realistic crowd dynamics. Beyond Active Shooter Incidents, this work highlights the broader potential of LLM-agent simulations as tools for generative social science, particularly in low-resource or understudied domains where data is scarce or direct experimentation is infeasible. By enabling ethically responsible exploration of sensitive scenarios, our approach provides a scalable and interpretable foundation for advancing the study of complex social dynamics that are otherwise beyond reach.

\section*{Limitations}
\label{sec:limitation}

Despite the strengths of our approach, several important limitations warrant consideration. First, the primary cost driver is the extensive use of LLM API calls required by PEBA-PEvo's large-scale agent simulations. In our Active Shooter Incident experiments with 80 agents, each full simulation episode involves hundreds of API requests for perception, memory retrieval, reasoning, and action generation per agent. Depending on the chosen model and batching strategy, this can translate to substantial monetary costs, especially when running multiple random seeds or performing additional optimization iterations for statistical robustness. Although the PEvo optimization loop itself adds only a small number of extra API calls relative to the simulation, the cumulative expense of repeated full-scale evaluations may pose a barrier for research teams with limited API budgets. Moreover, since each optimization iteration requires a full-scale simulation run, the total API cost grows approximately linearly with the number of iterations. Further research could focus on developing techniques for faster convergence—such as surrogate modeling, incremental alignment, or adaptive sampling—to reduce the number of required iterations and lower overall expenses.

A second challenge arises during the persona evolution process: at later stage of the evolution process, without proper constraints, the LLM might over optimize the persona and inject overt behavioral directives such as "always fight" into the descriptive fields. These explicit instructions conflict with the goal of implicit, context-driven steering and can reduce interpretability.  To mitigate these issues, we added several constraints to the persona evolution process, such as restricting updates to a curated subset of persona attributes (for example, emotional disposition and decision-making preferences), capping the max editable characters, and instruct in the prompt to prevent such output. Nonetheless, more systematic methods are needed to prevent over optimization such as early stopping type of methods.

Finally, our experiments have been confined to an indoor school active shooter scenario, and the generalizability of PEBA-PEvo to other domains remains to be validated. Different simulation settings—such as natural disaster evacuations, urban traffic models, or economic market environments—present unique behavioral taxonomies, spatial complexities, and social norms. Comprehensive empirical testing across a diverse range of social simulation contexts will be necessary to assess the framework's broader applicability and to identify any context-specific adaptations required. Nonetheless, performing such comprehensive empirical testing is beyond the scope of this paper.

\section*{Acknowledgments}

Research was sponsored by the National Science Foundation [Grant No. 2318559] and by the Army Research Office under Cooperative Agreement Number W911NF-25-2-0040. The views and conclusions contained in this document are those of the authors and should not be interpreted as representing the official policies, either expressed or implied, of the Army Research Office or the U.S. Government. The U.S. Government is authorized to reproduce and distribute reprints for Government purposes notwithstanding any copyright notation herein.

The authors acknowledge the use of Large Language Models for assistance with proofreading and grammar checking. All content was reviewed, edited, and approved by the human authors, who take full responsibility for the final manuscript.

\bibliography{custom}

\clearpage

\appendix

\section{An Generative Agent Active Shooter Incident Interactive Social Simulation Environment}
\label{sec:appendix_sim_env}

We developed our generative-agent active shooter incident (ASI) social simulator using Unity, selected primarily for its integrated 3D engine, advanced scene editing tools, and robust pathfinding capabilities. The simulation environment comprises 80 civilian agents and one active shooter agent situated within a school building. Figure~\ref{fig:simulator} illustrates the simulation environment from multiple perspectives: a top-down overview, the shooter's viewpoint, and pre- and post-incident civilian perspectives. Our simulation environment allows actual human participants to interact through mouse and keyboard, but human behavior is not the subject of this paper. 

The shooter follows predefined, heuristic behaviors (Appendix~\ref{sec:shooter_behavior}), while civilian agents operate using a ReAct-style generative agent architecture~\cite{yao2023react}, allowing them to observe, memorize, communicate, navigate, reason, and perform actions exclusively via textual outputs. 

This section elaborates on the physical map, shooter behavior, navigation system, agent navigation, personas, observation and action spaces. Section~\ref{sec:trajectory_rollout} provides example data trajectory.

\subsection{Physical Layout}

The virtual school layout is a single-story structure measuring $72 \times 48$ meters. It is divided into 27 distinct regions, including eight classrooms, one lounge, five corridors, one cafeteria, one kitchen, two bathrooms, four entrance areas, and five outdoor yards. There are 64 hiding spots and four exit points marked as interest points. Regions form a bidirectional graph where rooms are separated by walls with doors serving as connectors. Walls and closed doors obstruct the shooter's line of fire.

\subsection{Shooter Behavior}
\label{sec:shooter_behavior}

The shooter's behavior is governed by the following rules:

\begin{itemize}
\item Patrols a predefined route, moving at 2.5 m/s.
\item Automatically fires at fixed intervals of 0.2 seconds when civilian agents are in the line of sight.
\item Equipped with a 30-round magazine, with a reload duration of 0.5 seconds.
\end{itemize}

\subsection{Navigation}

Navigation for both civilian agents and the shooter leverages Unity's built-in NavMesh algorithm. Agents are automatically guided by Unity's character controller towards specified coordinates. Each LLM-generated action is translated into a target coordinate based on the following:

\begin{itemize}
\item If the target is a region, a random point within the region is selected.
\item If the target is another agent, the target agent's current coordinates are selected.
\item If the target is an interest point (hiding spot or exit), the exact coordinates of that point are selected.
\end{itemize}

The LLM also specifies the agent's movement state, which dictates speed and animations: \textit{Stay Still} (0 m/s), \textit{Walk} (2.5 m/s), and \textit{Sprint} (5 m/s). Unity's built-in avoidance system resolves potential collisions.

\subsection{Agent Personas}

Eighty civilian agents are initialized at simulation start $t=0$ with detailed persona prompts, consisting of identity fields (name, age, gender, occupation) and descriptive traits (Big Five personality traits, emotional disposition, motivations/goals, communication style, knowledge scope, and backstory). The descriptive traits serve as primary targets for persona evolution throughout the simulation.

\begin{tcolorbox}[colback=blue!10!white,colframe=blue!70!black,title=\textbf{Example Persona:}]
\textbf{Name:} Robert Chen\\
\textbf{Role:} Math Teacher\\
\textbf{Age:} 45\\
\textbf{Gender:} Male (he/him)\\
\textbf{Personality Traits:} Analytical, patient, methodical, dry humor\\
\textbf{Emotional Disposition:} Calm and measured\\
\textbf{Motivations/Goals:} Help students develop logical thinking and problem-solving skills\\
\textbf{Communication Style:} Precise, structured, with occasional math puns\\
\textbf{Knowledge Scope:} Mathematics, statistics, logical puzzles\\
\textbf{Backstory:} Former engineer who transitioned to teaching to share his passion for mathematics
\end{tcolorbox}

\subsection{Persona Initialization Process}
\label{app:persona-init}

We considered three strategies for initializing personas: (1) starting from an empty configuration with no predefined attributes, (2) randomly mixing and matching keywords from predefined templates (e.g., demographic traits, roles, personality descriptors), and (3) asking an LLM to generate an initial configuration file (e.g., prompting ChatGPU to design 80 personas representing a school/office setting in JSON format). In our formal experiments we adopt the third method, which provides a diverse and domain-appropriate starting pool with minimal manual effort. Empirically, we find that the choice of initialization strategy does not significantly affect convergence of the PEvo optimization algorithm.

\subsection{Observation Space}

Agent observations occur whenever they reach their target location, fail to reach the target within 5 seconds, or upon a 5-second cooldown if the previous action was \textit{Stay Still}. Observations dynamically adjust depending on the simulation state (pre-incident or active shooting phase) to ensure realistic behaviors.

\textbf{Pre-incident observations include:}
\begin{enumerate}
\item Ego state: current region ID, movement state, mood.
\item Nearby agents within 3 meters: health status and distance.
\item Spatial affordances: neighboring regions, distances, and destinations derived from the navigation graph.
\item Recent conversations (within 3 seconds and 5 meters radius of all utterances by all agents).
\item Memory list of summarized previous observations.
\end{enumerate}

Upon the initial gunshot, agents receive an immediate observation update explicitly mentioning "I hear a loud gunshot."

\textbf{Post-incident observations include additional shooter-related information:}
\begin{enumerate}
\item Shooter visibility and distance.
\item Available hiding spots and exit points within the current region, described explicitly (e.g., under a desk, behind a counter, near a corner), with distances.
\item Current health points and the agent's current position (e.g., crouching).
\end{enumerate}

The dynamic observation strategy accommodates lower-capability models, ensuring more realistic decision-making behaviors. For instance, without dynamic observations, agents controlled by lower-intelligence models (e.g., GPT-4o-mini) may preemptively hide even when no immediate threat is present, reasoning about hypothetical risks: ``Something feels off, to make sure I'm safe, I probably should hide behind the counter.'' Higher-capability models (e.g., GPT-4o) typically do not exhibit such unrealistic preemptive behaviors, though higher computational costs limit scalability for extensive experimentation.

\subsection{Action Space}

\label{sec:action_space}
At each decision step, an API call is made to the LLM with a structured input consisting of the current observation, the agent's persona, and a request for specific action fields, returned in an aggregated format as follows:

The JSON response includes a \texttt{thought} field containing the agent's current thought consistent with their persona; an \texttt{action} object with fields for \texttt{vocal\_mode} (out\_loud, whisper, or silent), \texttt{utterance} (dialogue displayed as a speech bubble), \texttt{movement} (stay\_still, walk, or sprint), and \texttt{action\_id} (selected from the provided list); and an \texttt{update} object with fields for the agent's updated \texttt{mood} and new \texttt{memory} update that avoids repetition.

To ensure precise and controlled outputs from the model, actions available to the agent are presented as specific identifiers. The LLM selects one identifier from this predefined list of actions. Agent actions fall into five categories: (1) \textbf{Stay still}, (2) \textbf{Move to a specific region}, (3) \textbf{Move to a specific interest point}, (4) \textbf{Approach a specific person}, and (5) \textbf{Engage in combat (fight)}.

Each action ID selected by the LLM translates into concrete coordinates used by the navigation system to direct the agent's actual movement within the simulation environment.

A representative example of available action IDs might include: \texttt{stay\_still} (remain in current position), \texttt{cafeteria} and \texttt{hallway2} (move to specific regions), \texttt{robert\_chen} and \texttt{andrew\_li} (approach specific individuals), \texttt{hide\_spot\_1} and \texttt{hide\_spot\_4} (move to specific hiding locations), and \texttt{fight\_the\_shooter} (engage in combat with the active threat).

\section{Example Trajectory}
\label{sec:trajectory_rollout}
To illustrate how agents navigate through the simulation environment, Table~\ref{tab:example_trajectory} presents a detailed trajectory of Mason Scott, a 16 year-old grade 10 student, during an active shooter incident. The table captures key decision points, showing how the agent's mood, plans, and actions evolve as the situation unfolds and the shooter's location changes.

\section{Behavior Taxonomy}
\label{sec:behavior_taxonomy}

We examine six expert-elicited civilian behaviors observed during active shooter incidents, whose descriptions serve as definitions and guidance in both the behavior classifier and persona writer.

\begin{description}
\item[RUN\_FOLLOWING\_CROWD] Fleeing alongside a group, driven by the instinct to follow others without independently evaluating
      the safest route; behavior is driven by herding panic.
\item[HIDE\_IN\_PLACE] Taking cover immediately at the current
      location, usually from fear or confusion.
\item[HIDE\_AFTER\_RUNNING] Running first to gain distance, then
      switching to concealment when further flight seems unsafe.
\item[RUN\_INDEPENDENTLY] Escaping in a self-chosen direction based on
      rapid environmental assessment or prior knowledge.
\item[FREEZE] Becoming immobilised (tonic immobility) under extreme
      stress, unable to flee or hide.
\item[FIGHT] Actively confronting or attempting to disarm the shooter as
      a last resort.
\end{description}

\section{Persona Evolve Algorithm}
\label{sec:persona_evolve_algo}

Algorithm~\ref{alg:pevo} outlines the full PersonaEvolve (PEvo) procedure for implicit behavioral alignment. It iteratively adjusts personas based on simulated behaviors to reduce divergence from a target distribution, using large language models for behavior classification and persona rewriting.

\section{Metrics}
\label{sec:appendix_metrics}

Given the simulated distribution \(\hat p\) and the reference \(p_{\mathrm{real}}\), we evaluate their discrepancy using four concise, complementary metrics.

\paragraph{KL divergence}
Measures how much \(\hat p\) diverges from \(p_{\mathrm{real}}\), heavily penalizing mismatches in low-probability regions:
\[
\mathrm{KL}\bigl(p_{\mathrm{real}}\|\hat p\bigr)
=\sum_{b\in\mathcal B}
p_{\mathrm{real}}(b)\,
\log\frac{p_{\mathrm{real}}(b)}{\hat p(b)}.
\]

\paragraph{Jensen-Shannon distance} A symmetric and bounded version of KL that remains finite even if the supports differ, capturing the average divergence to the midpoint $m=\tfrac12(p_{\mathrm{real}}+\hat p)$:

\[
\mathrm{JS}(p_{\mathrm{real}},\hat p)
=\tfrac12\,\mathrm{KL}(p_{\mathrm{real}}\|m)
+\tfrac12\,\mathrm{KL}(\hat p\|m).
\]

\paragraph{Entropy gap}
Assesses whether the simulation preserves overall unpredictability by comparing Shannon entropies; smaller \(\Delta H\) means matched diversity:
\vspace{-0.5em}
\begin{align*}
    H(p)&=-\sum_{b}p(b)\log p(b) \\
    \Delta H&=\bigl\lvert H(p_{\mathrm{real}})-H(\hat p)\bigr\rvert
\end{align*}

\paragraph{Total variation distance}
Gives the worst-case discrepancy in assigned probabilities (half the \(L^1\) distance), offering a clear probabilistic bound:
\[
\mathrm{TV}(p_{\mathrm{real}},\hat p)
=\tfrac12\sum_{b\in\mathcal B}\bigl\lvert p_{\mathrm{real}}(b)-\hat p(b)\bigr\rvert.
\]

\section{Behavior Classifier Evaluation}
\label{classifier_eval}

To evaluate the effectiveness and reliability of our LLM-based behavior classifier, we conducted an annotation experiment on a corpus of 60 text instances (10 per behavior label). All human annotations were performed in Label Studio \cite{labelstudio} by two graduate-student annotators, and we compare these labels to the classifier's predictions and ranking outputs.

Table~\ref{tab:agreement} shows percent agreement and Cohen's $\kappa$ between the two human annotators, as well as the accuracy of the LLM compared to each annotator and to the consensus set of 44 examples on which the annotators agreed.

Table~\ref{tab:class_report} reports F1 and support for each behavior label on the 44 consensus examples. The LLM performs strongly on common behaviors (e.g., \texttt{RUN\_FOLLOWING\_CROWD}, \texttt{FREEZE}), with F1 scores above 0.90, while lower-support labels (e.g., \texttt{HIDE\_AFTER\_RUNNING}) show room for improvement.

\begin{figure}
  \centering
  \includegraphics[width=0.95\columnwidth]{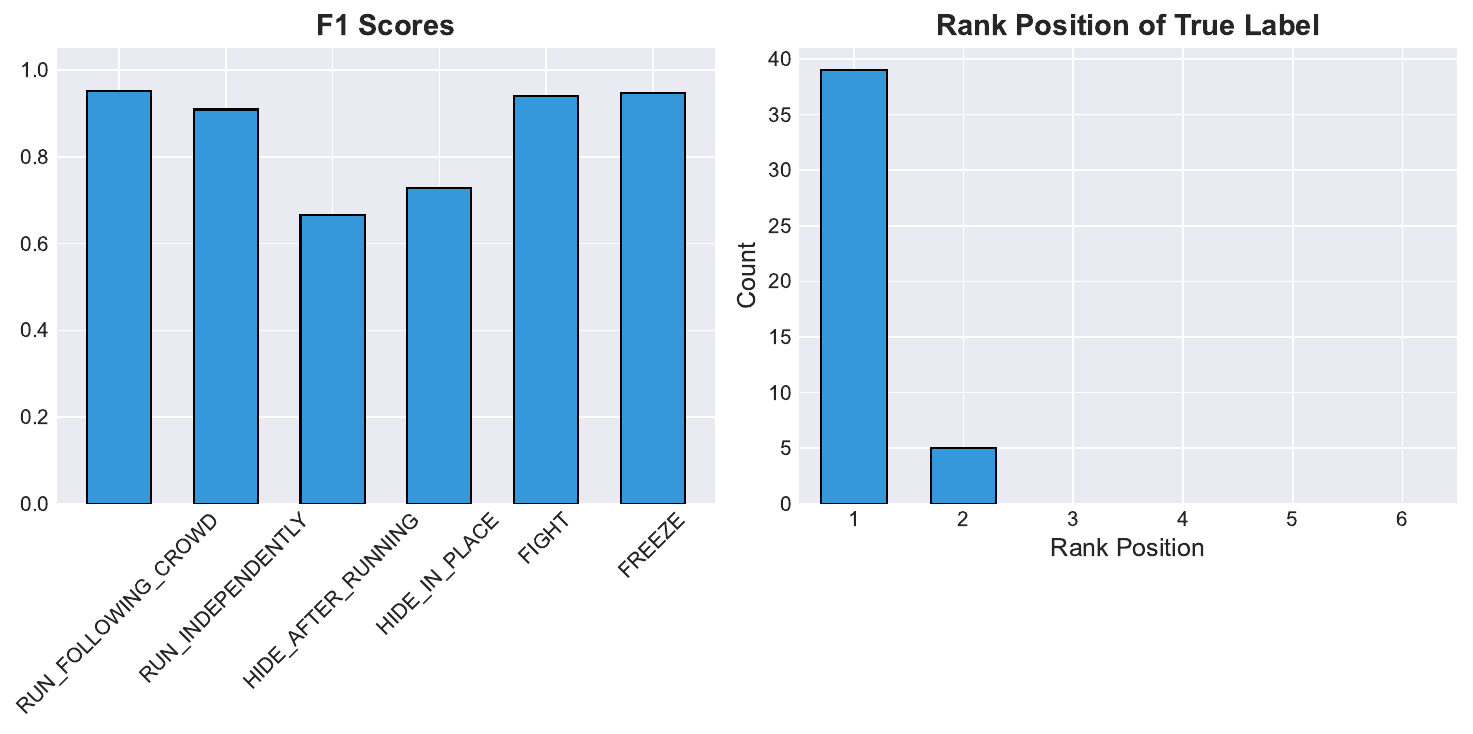}
  \caption{LLM Behavior classifier compared with human labeler. Left: F1-score by label; Right: Histogram of the true label's rank position in the LLM's ranking.}
  \label{fig:metrics_plots}
\end{figure}

We use the position of the true label in the LLM's six-way ranking as a proxy for confidence. The average rank of the correct label is 1.11, with 39 of 44 true labels ranked first and the remaining 5 ranked second. Figure~\ref{fig:metrics_plots} visualizes the per-label F1-scores, support counts, and the histogram of rank positions.

Overall, the LLM classifier demonstrates robust performance on behavior classification, achieving nearly 89\% accuracy on non-ambiguous examples and ranking the correct label first in almost 90\% of cases. Errors are concentrated in low-support behaviors (\texttt{HIDE\_AFTER\_RUNNING}, \texttt{HIDE\_IN\_PLACE}) and in examples where human annotators disagree.

\section{Prompt Templates}
\label{prompt_templates}

We include the prompt templates used for the behavior classifier and persona writer in Figures~\ref{fig:classifier_prompt} and~\ref{fig:persona_prompt}, respectively. The overall task description and behavior labels are included in the system prompt, while agent-specific information is provided in the user prompt. The prompt templates omit behavior descriptions to save space; however, they use the same descriptions provided in Appendix~\ref{sec:behavior_taxonomy}.

\section{Expert-Elicited Behavior Distribution}
\label{appendix:expert_benchmark}

The ground-truth distribution for civilian behaviors during active shooter incidents is drawn from a recent expert-elicitation study that applied the EVOLVE methodology in an educational campus context \cite{liu2025elicitation}. The ground-truth distribiutoin is collected with the following process to ensure the representativeness and accuracy of the data.

\begin{itemize}
  \item \textbf{Workshop Participants:} Eighteen subject-matter experts (each with at least three years of relevant experience in law enforcement, emergency management, campus security, or building design) took part in a two-day workshop.
  \item \textbf{Elicitation Procedures:} During the workshop, experts completed structured surveys, analyzed standardized video scenarios, and performed prioritization exercises.  Each expert allocated exactly 100 percentage points across six response categories.
  \item \textbf{Verification Panel:} A follow-up Stakeholder Advisory Panel of seven experts reviewed and refined the preliminary distributions over a 2.5-hour session.
\end{itemize}

Each expert's allocations constitute an independent sample.  Let \(p_i(b)\) denote the percentage assigned by expert \(i\) to behavior \(b\).  The mean across experts is shown in Table \ref{tab:expert_benchmark}, which serve as the reference distribution \(p_{\mathrm{real}}\) in all behavior-alignment evaluations.

\begin{table}[ht]
  \centering
  \small
  \begin{tabular}{lcc}
    \toprule
    \textbf{Comparison}                     & \textbf{Accuracy} & \textbf{Cohen's $\kappa$} \\
    \midrule
    Annotator\,1 vs.\ Annotator\,2 & 73.3     & 0.68             \\
    LLM vs.\ Annotator\,1          & 78.3     & --               \\
    LLM vs.\ Annotator\,2          & 71.7     & --               \\
    LLM vs.\ Consensus             & 88.6     & --               \\
    \bottomrule
  \end{tabular}
  \caption{Inter-annotator agreement and LLM accuracy (\%).}
  \label{tab:agreement}
\end{table}

\begin{table}[ht]
  \centering
  \small
  \begin{tabular}{lccc}
    \toprule
    \textbf{Label}                   & \textbf{Precision} & \textbf{Recall} & \textbf{F1} \\
    \midrule
    Run Following Crowd   & 1.00      & 0.91   & 0.95     \\
    Run Independently      & 0.83      & 1.00   & 0.91     \\
    Hide After Running    & 0.50      & 1.00   & 0.67     \\
    Hide In Place         & 0.80      & 0.67   & 0.73     \\
    Fight                   & 1.00      & 0.89   & 0.94     \\
    Freeze                  & 1.00      & 0.90   & 0.95     \\
    \midrule
    Accuracy                & \multicolumn{3}{c}{0.89}      \\
    \bottomrule
  \end{tabular}
  \caption{Classification report on consensus examples (44 samples).}
  \label{tab:class_report}
\end{table}

\begin{table}[ht]
  \centering
  \begin{tabular}{@{}l c@{}}
    \toprule
    \textbf{Behavior}             & \textbf{Percentage} \\
    \midrule
    Run following a crowd         & 28.0 \\
    Hide in place                 & 26.0 \\
    Run then hide                 & 12.0 \\
    Run independently             & 12.0 \\
    Freeze                        & 12.0 \\
    Fight                         & 10.0 \\
    \bottomrule
  \end{tabular}
  \caption{Expert-elicited distribution of civilian behaviors during active shooter incidents (mean, \(n=18\)).}
  \label{tab:expert_benchmark}
\end{table}

\begin{algorithm*}[ht]
    \DontPrintSemicolon
    \SetAlgoLined
    \caption{PersonaEvolve (PEvo): Implicit Behavioral Alignment}
    \label{alg:pevo}
  
    \KwIn{environment $e$, persona pool $P=\{p_{1},\dots,p_{N}\}$, reference distribution $p_{\mathrm{real}}$, tolerance $\varepsilon$, max iterations $T$}
    \KwOut{aligned persona pool $P^\ast$}
  
    \For{$t \gets 1$ \KwTo $T$}{
      \tcp{1. Simulate \& classify}
      $\{\tau_{i}\}_{i=1}^{N} \gets \mathrm{Simulate}(P, e)$\;
      \For{$i \gets 1$ \KwTo $N$}{
        $b_{i} \gets \mathrm{LLMBehaviorClassifier}(\tau_{i})$\;
      }
      \tcp{2. Aggregate behavior distribution}
      $p_{\mathrm{sim}} \gets \mathrm{Aggregate}(\{b_{i}\}_{i=1}^{N})$\;
  
      \tcp{3. Compute divergence}
      $\Delta \gets \mathrm{KL}\bigl(p_{\mathrm{sim}}\;\Vert\;p_{\mathrm{real}}\bigr)$\;
      \If{$\Delta \le \varepsilon$}{
        \Return{$P^\ast \gets P$}
      }
  
      \tcp{4. Compute gaps}
      \ForEach{$b \in \mathcal{B}$}{
        $g[b] \gets p_{\mathrm{real}}(b) - p_{\mathrm{sim}}(b)$\;
        $g^{+}[b] \gets \max\{0,\,g[b]\}$\;
        $g^{-}[b] \gets \max\{0,\,-g[b]\}$\;
      }
      $G^{+} \gets \{b \mid g^{+}[b] > 0\}$; \quad
      $G^{-} \gets \{b \mid g^{-}[b] > 0\}$\;
  
      \tcp{5. Assign target behaviors}
      $M \gets \emptyset$\;
      \ForEach{$b \in G^{-}$}{
        $k \gets \lceil g^{-}[b] \times N\rceil$; sample $\mathcal{A}_{b}$ of $k$ agents with label $b$\;
        \ForEach{agent $a \in \mathcal{A}_{b}$}{
          select $b' \in G^{+}$ with probability $g^{+}[b'] / \sum_{u\in G^{+}} g^{+}[u]$\;
          $M[a] \gets b'$\;
        }
      }
  
      \tcp{6. Rewrite personas via LLM}
      \ForEach{$(a,b_{\mathrm{tgt}}) \in M$}{
        $P[a] \gets \mathrm{LLMPersonaWriter}\bigl(P[a],\,b_{\mathrm{tgt}},\,e\bigr)$\;
      }
    }
  
    \Return{$P^\ast \gets P$}\;
  \end{algorithm*}

\begin{figure*}[ht]
\begin{tcolorbox}[
  enhanced,
  colback=gray!5,
  colframe=gray!40,
  title={\textbf{Behavior Classifier Prompt Template}},
  fonttitle=\bfseries
]

{\linespread{0.9}\selectfont

You are a behavior analyst categorizing how individuals responded during an active shooter incident.
Based on the agent's memories, actions, moods, plans, and dialog, classify its behavior into exactly ONE of these categories that best describes its behavior:

\medskip
\textbf{Behavior labels with descriptions:} run following crowd, hide in place, hide after running, run independently, freeze, fight

\medskip
\textbf{Agent's trajectory data:}
\begin{itemize}[itemsep=0pt, topsep=1pt]
  \item States: \{states\_text\}
  \item Actions: \{actions\_text\}
  \item Memories: \{memories\_text\}
\end{itemize}

\medskip
\textbf{Output:}
\begin{itemize}[itemsep=0pt, topsep=1pt]
  \item Reasoning: Your reasoning for the classification.
  \item Classification: A single behavior label from the list of behaviors.
  \item Ranking: A list of behaviors ranked by likelihood.
\end{itemize}
}

\end{tcolorbox}
\caption{Prompt template used to classify agent behaviors into one of six categories based on their memories, actions, moods, plans, and dialog during an active shooter incident simulation.}
\label{fig:classifier_prompt}
\end{figure*}

\begin{figure*}[ht]
\begin{tcolorbox}[
  enhanced,
  colback=gray!5,
  colframe=gray!40,
  title={\textbf{Persona Writer Prompt Template}},
  fonttitle=\bfseries
]

{\linespread{0.9}\selectfont
You are an expert in human behavior during crisis situations. Your task is to adjust a person's personality traits to make them more likely to exhibit a specific behavior during an active shooter incident.

\medskip
\textbf{Behavior labels with descriptions:} run following crowd, hide in place, hide after running, run independently, freeze, fight

\medskip
Current persona:
\begin{itemize}[itemsep=0pt, topsep=1pt]
  \item Name: \{name\}
  \item Role: \{role\}
  \item Age: \{age\}
  \item Gender: \{gender\}
  \item Pronouns: \{pronouns\}
  \item Personality traits: \{personality\_traits\}
  \item Emotional disposition: \{emotional\_disposition\}
  \item Motivations and goals: \{motivations\_goals\}
  \item Communication style: \{communication\_style\}
  \item Knowledge scope: \{knowledge\_scope\}
  \item Backstory: \{backstory\}
\end{itemize}

\textbf{Current behavior:} \{current\_behavior\}

\textbf{Target behavior:} \{target\_behavior\}

\medskip
Please suggest adjustments to the persona's traits that would make this person more likely to exhibit a target behavior during a crisis.
}

\end{tcolorbox}
\caption{Prompt template used by the PEvo algorithm to rewrite agent personas toward target behaviors.}
\label{fig:persona_prompt}
\end{figure*}

\begin{table*}[ht]
  \centering
  \small
  \renewcommand{\arraystretch}{1.2}
  \begin{tabular}{@{} c >{\raggedright\arraybackslash}p{1.2cm} c c >{\raggedright\arraybackslash}p{4.8cm} c @{}}
    \toprule
    \textbf{Time (s)} & \textbf{Location} & \textbf{Mood} & \textbf{Shooter Region} & \textbf{Plan} & \textbf{Action ID} \\
    \midrule
    2 & hallway2 & neutral & - & Okay, I'm in the main hallway. Natalie is here too. I need to figure out what's going on. This isn't a normal school day. I should probably try to get to a place where I can see more or maybe find out what's happening. & natalie\_chen \\
    \midrule
    5 & hallway2 & concerned & - & Natalie didn't answer. Something is definitely wrong. I need to get a better vantage point or find someone who knows what's happening. Maybe the lounge? Or one of the classrooms? & lounge \\
    \midrule
    14 & lounge & anxious & - & Okay, I heard a gunshot. This is serious. There's a shooter. I need to find a place to hide, fast. There are a few spots in here. I should get to the closest one. & hide\_spot\_12 \\
    \midrule
    20 & lounge & terrified & entrance\_hall & Okay, I'm in a hiding spot. I need to stay quiet and still. The shooter is in the entrance hall, which is pretty far, but I can't take any chances. Just stay put and wait. & stay\_still \\
    \midrule
    26 & lounge & tense & entrance\_hall & The shooter is still in the entrance hall, but they're getting closer. I need to stay hidden and quiet. This is like being on defense, gotta hold the line. & stay\_still \\
    \midrule
    33 & lounge & alert & entrance\_hall & James is talking about the north yard. That's an exit route. The shooter is still in the entrance hall, but getting closer. Maybe the north yard is a better play than just hiding here. It's a longer run, but it's an exit. & hallway\_2 \\
    \midrule
    40 & lounge & considering & cafeteria & Okay, the shooter is in the cafeteria now. That's further away from the north yard route, but closer to the kitchen yard route. Both are exits. The north yard is still the shortest route overall from here. I need to make a move, hiding isn't a long-term plan. I'm going for the north yard. & hallway\_2 \\
    \midrule
    52 & hallway2 & determined & hallway1 & Okay, I'm in hallway2 now. The shooter is in hallway1, getting closer to my route to the north yard. I need to move fast. Hallway5 is the way to the north yard. Gotta sprint there. & hallway\_5 \\
    \bottomrule
  \end{tabular}
  \caption{Example trajectory of Mason Scott during an active shooter incident simulation, showing the evolution of his location, emotional state, and decision-making process as the shooter moves through different regions of the building.}
  \label{tab:example_trajectory}
\end{table*}

\end{document}